\pdfoutput=1
\documentclass[letterpaper, 10 pt, conference]{IEEEtran}  

\IEEEoverridecommandlockouts                              

\usepackage[bookmarks=false]{hyperref}
\usepackage{graphics} 
\usepackage{epsfig} 
\usepackage{mathptmx} 
\usepackage{times} 
\usepackage{amsmath} 
\usepackage{amssymb}  

\usepackage[T1]{fontenc}
\usepackage{amsfonts}
\usepackage{booktabs}
\usepackage{siunitx}
\usepackage{caption} 
\usepackage{booktabs}

\usepackage{array}    

\usepackage{graphicx}
\usepackage{algorithm, algpseudocode}
\usepackage{float}
\usepackage[percent]{overpic}
\usepackage{caption}
\DeclareGraphicsExtensions{.jpeg,.png}

\usepackage[font=small,skip=0pt]{subcaption}

\usepackage{array, tabularx, makecell, booktabs}%

\usepackage{tikz}

\usepackage{geometry}
\usepackage{comment}

\usepackage{hyperref}
\hypersetup{
    colorlinks=true,
    linktoc=all,
    citecolor=black,
    filecolor=black,
    linkcolor=blue,
    urlcolor=black
}

\usepackage{footmisc}

\usepackage{xcolor}

\title{\LARGE \bf
Arena 4.0: A Comprehensive ROS2 Development and Benchmarking Platform for Human-centric Navigation Using Generative-Model-based Environment Generation
}

\author{ 
    Volodymyr Shcherbyna$^{1,*}$, 
    Linh K{\"a}stner$^{1,2,*}$,
    Diego Diaz$^{1}$,  \\
    Huu Giang Nguyen$^{1}$,
    Maximilian Ho-Kyoung Schreff$^{1}$, 
    Tim Seeger$^{1}$,\\
    Jonas Kreutz$^{1}$,
    Ahmed Martban$^{1}$,
    
    Huajian Zeng$^{3}$,
    Harold Soh$^{2}$\\
    {\small $^{*}$Equal Contribution}\\
    {\small $^{1}$Technical University Berlin (TUB), Germany} \\
    {\small $^{2}$National University of Singapore (NUS), Singapore} \\
    {\small $^{3}$Technical University Munich (TUM), Germany}
}

\usepackage{makecell}

\begin{document}

\maketitle
\thispagestyle{empty}
\pagestyle{empty}

\begingroup
\renewcommand\thefootnote{}
\footnotetext{\hspace*{-2.2em}This research / project is supported by A*STAR under its National Robotics Programme (NRP) (Award M23NBK0053). This research is supported in part by the National Research Foundation (NRF), Singapore and DSO National Laboratories under the AI Singapore Program (Award Number: AISG2-RP-2020-017).}
\endgroup


\begin{abstract}
\noindent
Building upon the foundations laid by our previous work, this paper introduces Arena 4.0, a significant advancement of Arena 3.0 \cite{arena3}, Arena-Bench \cite{arena:bench}, Arena 1.0 \cite{arena:rosnav}, and Arena 2.0 \cite{arena-rosnav-2.0}. Arena 4.0 provides three main novel contributions: 1) a generative-model-based world and scenario generation approach using large language models (LLMs) and diffusion models, to dynamically generate complex, human-centric environments from text prompts or 2D floorplans that can be used for development and benchmarking of social navigation strategies. 2) A comprehensive 3D model database which can be extended with 3D assets and semantically linked and annotated using a variety of metrics for dynamic spawning and arrangements inside 3D worlds. 3) The complete migration towards ROS 2, which ensures operation with state-of-the-art hardware and functionalities for improved navigation, usability, and simplified transfer towards real robots. We evaluated the platforms performance through a comprehensive user study and its world generation capabilities for benchmarking demonstrating significant improvements in usability and efficiency compared to previous versions. Arena 4.0 is openly available at \href{https://github.com/Arena-Rosnav}{\color{blue}https://github.com/Arena-Rosnav}.

\end{abstract}

\section{Introduction}
\noindent As the field of autonomous robots evolves, the demand for robots capable of navigating and interacting within human-centric environments continues to grow, particularly in areas such as healthcare, logistics, and assistive robotics \cite{mirsky2024conflict}. To develop and test social navigation approaches that operate these robots, realistic and comprehensive simulation platforms are essential. However, the transition from simulation to real-world applications (sim2real) still remains a substantial bottleneck, with many existing platforms still reliant on outdated software such as ROS1, making the transition difficult and exacerbating the sim2real gap \cite{singamaneni2024survey}.
\begin{figure}[!htp]
    \centering
    \includegraphics[width=0.99\linewidth]{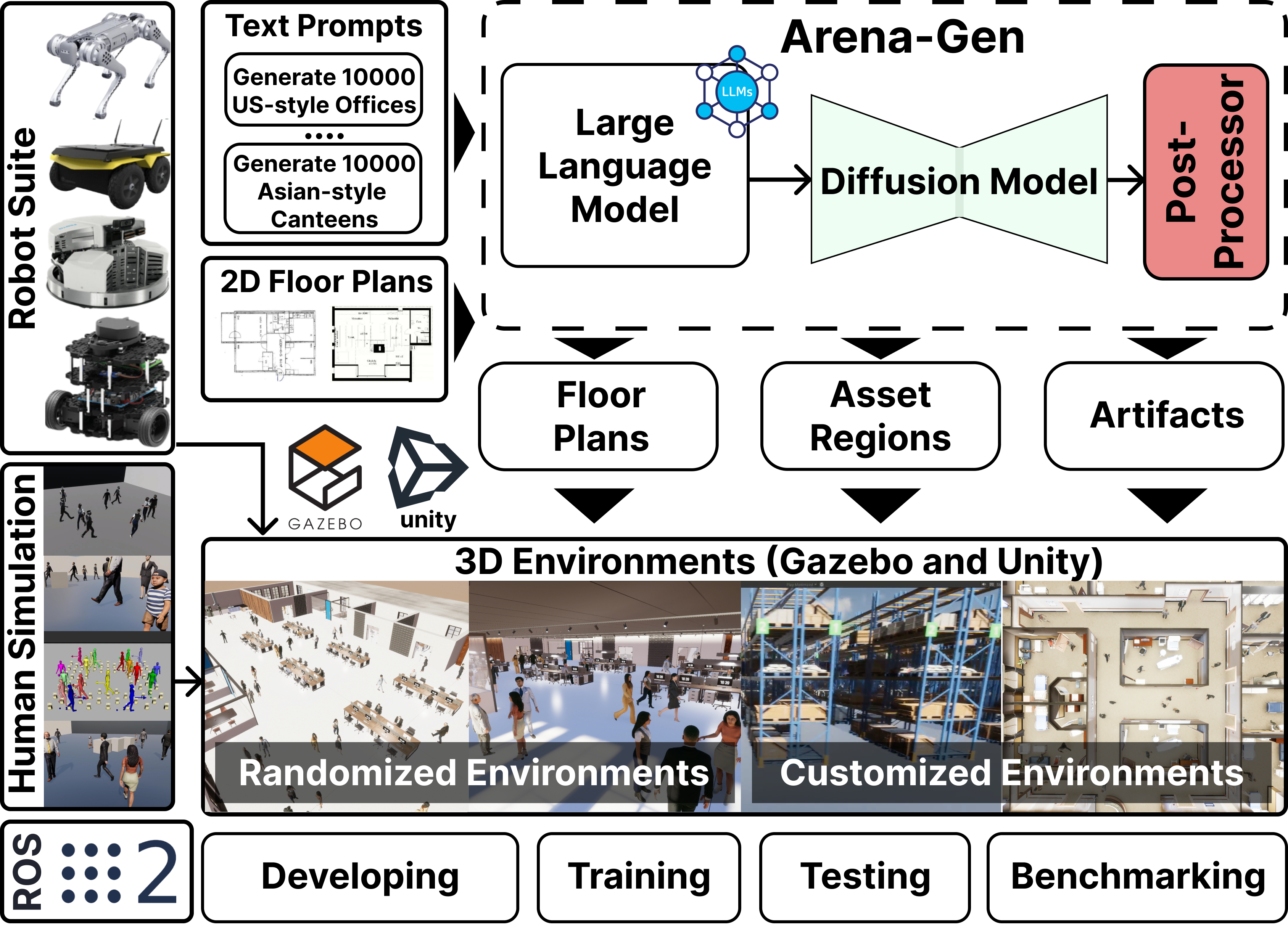}
    \caption{Arena 4.0 is a software stack of tools to develop and benchmark social navigation algorithms. It fully operates under ROS2 making it compatible with state of the art hardware and eases production level development. We introduce a novel dynamic world generator using generative models, indicated as Arena-gen. Together with various other usability enhancements, Arena 4.0 is highly suitable for unified benchmarking and development. It was selected as the host platform for the \href{https://socialnav2025.pages.dev/organization}{\color{blue}SocialNav2025 competition}.}
    \label{intro}
\end{figure}
\\\noindent
Furthermore, despite recent advancements and the growing demand for social robotics, many existing works remain limited by overly specific implementations, lack of reproducibility, and fragmentation across different simulation platforms, which hinders broader adoption and industrial deployment \cite{singamaneni2024survey, mirsky2024conflict, francis2023principles}. As an effort to address these existing issues, in our previous work, Arena 3.0, we provided a unified platform to develop and test social navigation approaches in realistic social settings by incorporating sophisticated human behavior models and social interaction patterns, focusing on the complexities of social navigation \cite{arena3}. 
\\\noindent Building on these foundations, Arena 4 introduces a fully ROS2-based development and benchmarking platform to not only facilitate unified benchmarking and testing of social navigation approaches but also to improve the ease of transitioning from simulation to real-world robotic systems. The adoption of ROS2 is an important step towards closing the sim2real gap. The platform also integrates state-of-the-art generative AI techniques such as large language models (LLMs) and diffusion models to dynamically generate complex environments for development and benchmarking. The main contributions of this work are as follows:
\begin{itemize}
    \item The full transition of the platform to ROS2 with a number of new functionalities to ease production level development;
    \item A world and scenario generation pipeline using generative models;
    \item A 3D models database with semantically linked assets for dynamic and customizable world and scenario generation 
    \item Additional improvements such extended human behavior simulation with HuNavSim extension with social states for scenario generation, and enhanced usability features such as one click installer, user interfaces for creation and execution of processes, documentation, and tutorials;.
\end{itemize}

\color{black}


\section{Related Works}
\noindent
This work builds on our previous platforms Arena-Bench \cite{arena:bench}, Arena 1.0 \cite{arena:rosnav}, Arena 2.0 \cite{arena2}, and Arena 3.0 \cite{arena3} extended them with extended world and scenario generation, benchmarking, and usability functionalities. Furthermore it is fully migrated to ROS2 providing all its functions. This latest version emphasizes dynamic generation of social environments as well as addressing the needs for production level and operating with the latest industrial available robots. Furthermore, the platform employs a more stable version and adds capabilities for benchmarking and competitions due to the new dynamic world and scenario generation aiming for the platform to serve as a host platform for international competitions. 
\\\noindent Robot navigation and obstacle avoidance in dynamic environments have been a well studied problem in robotics with a diverse range of research works employing both classic control theory as well as learning based approaches using reinforcement or imitation learning \cite{everett2018motion}, \cite{dugas2020navrep}, \cite{chen2019crowd}, \cite{chen2017socially}. More recently, a number of approaches utilized large language or diffusion models to incorporate socially aware navigation into the system \cite{song2024socially, sridhar2024nomad, hirose2023sacson, zu2024language, singamaneni2021human, bae2024sit}. However, a recent study by Francis et al. \cite{francis2023principles} on principles and guidelines for social navigation research found out that comprehensive simulation platforms for navigation benchmarking were still lacking or limited in functionality. Similar conclusions were drawn by a recent studies by Singamaneni et al. \cite{singamaneni2024survey} and Mirsky et al. \cite{mirsky2024conflict}
\\\noindent
Social navigation platforms such as SEAN 1.0 and 2.0 by Tsoi et al. \cite{tsoi2020sean, tsoi2022sean} provided valuable insights but were limited in terms of environment variety and simulation capabilities. For instance, Habitat 3.0 \cite{puig2023habitat} focuses solely on home environments whereas other platforms such as HuNavSim \cite{perezhunavsim} or SocialGym 2.0 \cite{sprague2023socialgym} are limited to specific simulators such as Gazebo or customized 2D simulators. Other benchmarking platforms such as Bench-MR by Heiden et al. \cite{bench_mr}, Robobench by Weisz et al. \cite{robobench}, CommonRoad by Althoff et al. \cite{althoff2017commonroad}, and the Benchmarking suite by Moll et al. \cite{moll2015benchmarking} contributed significantly to static environment navigation but did not address dynamic human-robot interaction complexities. Benchmarks that include dynamic pedestrian movements such as MRPB 1.0 by Wen et al. \cite{mrpb}, DynaBARN by Nair et al. \cite{nair2022dynabarn} or CoNav by Li et al. \cite{li2024conav} made strides in social navigation but faced limitations in scenarios, robot variety, or simulation environments.
\\\noindent
Our proposed platform addresses aforementioned gaps and aspire to provide a platform with comprehensive and dynamic world and scenario generation capabilities, a full transition to ROS2, as well as complete abstraction making it executable on multiple ubiquitous simulators, thus making it an ideal platform for de bencvelopment and benchmarking purposes.
\begin{figure*}[!htp]
    \centering
    \includegraphics[width=0.99\linewidth]{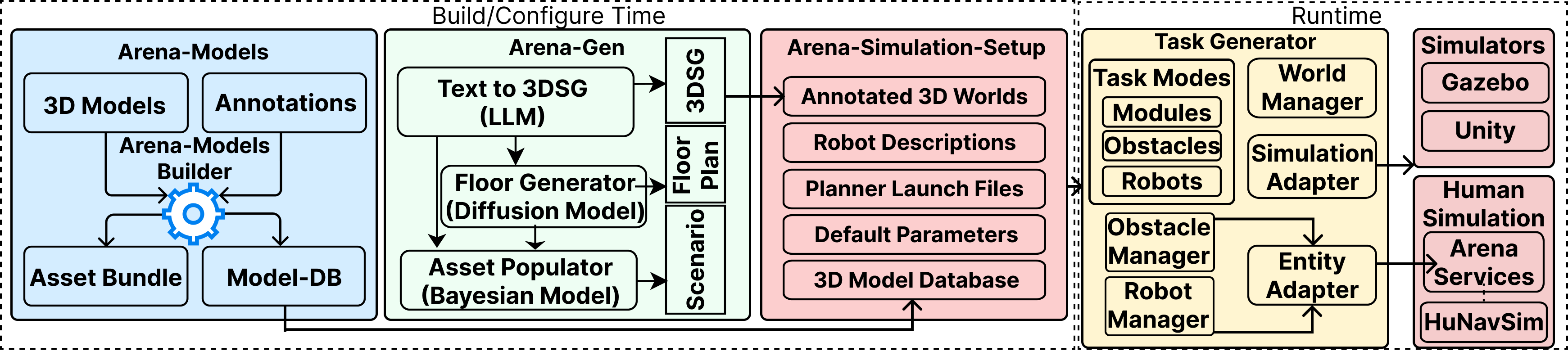}
    \caption{System Design of Arena 4.0. One of the main contributions is the improved dynamic world generation module, indicated as Arena-gen, using generative models and a language model based interface, which is accompanied by a web-based interface allowing users to customize world generation processes. Closely related, the Arena-Models module, which is a database that allows users to add and semantically link 3D models and assets for the system to plan for and spawn them in a dynamic manner. The Simulation-Setup and Task Generator modules from previous versions have been adapted and extended to take the generated worlds and semantic information as input and generate scenarios and tasks, now additionally based on scene graphs and room segmentations. As already introduced in previous works, our system design is completely abstracted and can thus be operated on multiple different simulators. The whole system has also been migrated to ROS2.}
    \label{img:a4-sd}
\end{figure*}

\section{Overview of Arena 4.0}
\noindent Similar to our previous works, Arena 4.0 is designed as a platform software stack of modules to facilitate the development and benchmarking of social navigation methods. 
\autoref{img:a4-sd} outlines the current version's modules. Arena 4.0 introduces several new key features:
\\\noindent
\textbf{Arena-gen.} One of the main novel contributions is an improved world generation module \emph{Arena-gen}, which employs generative models and a language-based interface, accompanied by a web-based interface, to allow users to customize world generation processes for an array of diverse applications. 
\\\noindent
\textbf{Arena-Models.} Closely related is another essential contribution, the \emph{Arena-Models} database that allows users to add and semantically link 3D models and assets for the system to plan for and spawn them in a dynamic manner. For creating and editing semantic annotations, we also provide an intuitive database explorer GUI. The Task generator and Simulation-Setup have been adapted and extended to receive those input and generate dynamic worlds and scenarios from the additional scene graph and room segmentation annotations. In the same manner, the respective simulators Gazebo and Unity have been adapted to reflect that information where applicable. 
\\\noindent
\textbf{ROS2 migration.} Another significant contribution is the complete migration to ROS2 to facilitate and simplify deployment in industrial production setups and robots, making newer functions accessible that are critical for production site robots, such as security, communication protocols, and improved navigation functionalities. This migration to ROS2 includes the navigation stack, which now operates on the Nav2 package, making it more accessible for newer robots that run exclusively on ROS2, such as Unitree's quadrupeds. It is important to note that the ROS1 version is still supported, and the possibility to run both in parallel is maintained until the end of the ROS1 lifecycle next year.
\\\noindent
\textbf{Human Behavior Models.} Finally, human behaviors have been optimized, extended, and refined. Specifically, we incorporated the open-source framework HuNavSim and extended it with more human-human and human-object interactions. In addition to these three core contributions, Arena 4.0 also adds a number of other miscellaneous functions such as new robots, planners, an improved training pipeline, a web application, or a comprehensive installer. In the following, each module will be described in more detail.

\subsection{Arena-gen: Generating Distinct Environments using Generative Models}
\noindent
The system design of the Arena-gen submodule is illustrated in \autoref{img:a4-sd}. It includes the pipeline for generating environments. Whereas in Arena 3, only three distinct randomized environment types could be generated, Arena Gen allows for the generation of a wide variety of indoor environments using language descriptions and diffusion models. 

\begin{figure*}[!htp]
    \centering
    \includegraphics[width=0.99\linewidth]{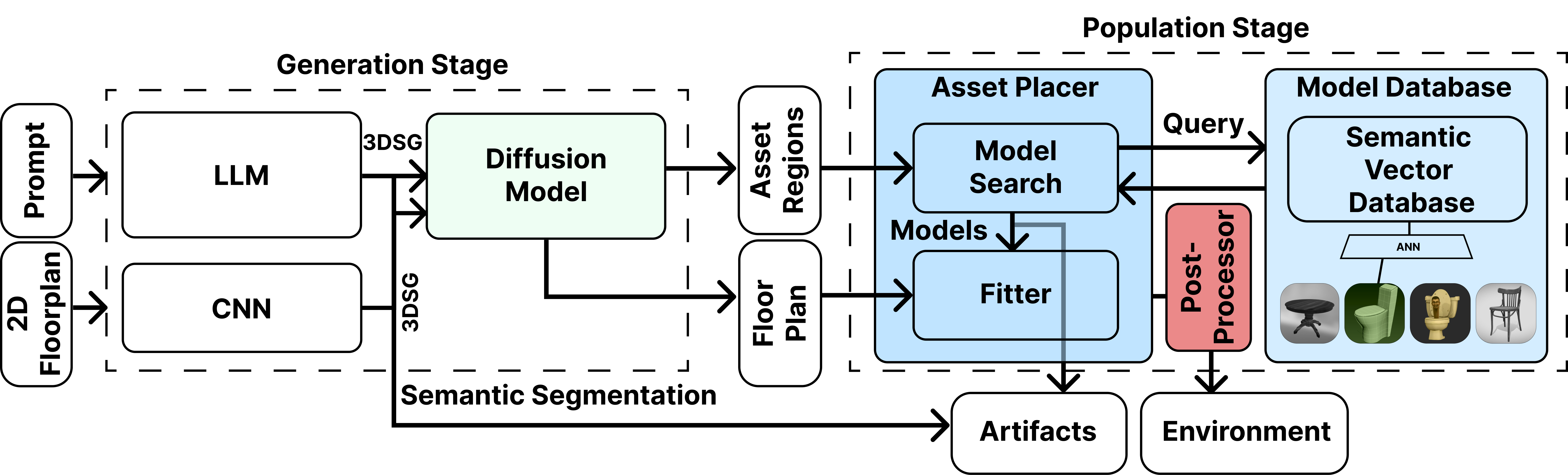}
    \caption{Data flow of the \emph{Generation Stage} and the \emph{Population Stage}. The Generation Stage combines multiple SotA technologies to process text inputs into a floor plan image and room asset locations. 3DSGs are used as an intermediate data structure to divide the problem into a text transformation task solvable by an LLM, and a graph transformation task solvable by a spatial GNN. The \emph{Population Stage} populates the floor plan's asset zones with 3D models by employing the \emph{Asset Placer}. A pre-built semantic vector \emph{Model Database} is queried for a related model, which is arranged into the zone by a \emph{Fitter} algorithm. After a final post-processing step, the end result is a finished environment consisting of 3D walls and models.}
    \label{img:gen-dataflow}
\end{figure*}

\autoref{img:gen-dataflow} illustrates the system design. Our generation process consists of a 2-stage pipeline, consisting of a Generation and a Population stage. The Generation stage uses an LLM that transforms natural language prompts into a a machine-readable graph. This graph is used in GNN inference to produce (1) an annotated floor plan image, and (2) asset regions placed within the rooms.
\\\noindent
During the Population Stage, the space of the assets is filled with models from a semantic model database. The floor plan, together with all placed models, is transformed into a final 3D environment that can be loaded by a simulator. Subdividing the process in this manner allows us to exploit the full extent of the expressiveness of the generative model during generation and then solving the model-populating sub-problem using more modular semi-classical approaches. 

\subsubsection{3D Scene Graph}
\noindent We introduce a variant of the 3DSG \cite{armeni20193d} as an intermediate format between the LLM and GNN components. The 3DSG format is a 2-level hierarchical graph that contains the room layout with room-room connections (upper half), and room-asset relationships (lower half) with asset descriptions. Assets are defined by a natural language description of the object, size, and color -- which the LLM extracts and supplements from user input. Preserving natural language descriptions throughout the pipeline allows us to decouple the embedding and thus semantics of each stage, making feature extraction fit each purpose instead of the data. 

\subsubsection{Dataset}
\noindent Our dataset consists of (3DSG, floor plan image) pairs that are used to train our GNN. The samples are based on the \emph{CubiCasa5K} \cite{kalervo2019cubicasa5k} dataset, which contains almost 5000 suitable real-world floor plan scans. The provided CubiCasa5K data processing detects the room and object segmentations using pre-trained CNNs, which we use as a base for our own data processing. We extract the 3DSG by (1) filtering and re-categorizing assets, (2) classifying doorways as either inter-room or external, (3) building a room connectivity matrix from inter-room doorways, (4) assigning the remaining assets to the sub-graph of the containing room. The result is a hierarchical graph in our 3DSG format paired with a cleaned floor plan image for direct insertion of our dataset.

\subsection{Model Database}
\noindent Another contribution of our work is the provision of a model database \emph{Arena-Models}, which is a semantic database of obstacle and human 3D models that can be spawned in the simulation.
We provide a workflow and builder scripts to process a directory containing models and annotations into 1) a prompt-queriable vector database of model IDs, 2) a simulator-specific Asset Bundle that stores the 3D models in a single compressed file.
End users can use our Python API to query the database and retrieve models, both as Unity prefabs and as exported COLLADA files for use in other simulators.
\\\noindent
Compared to the classic solution of loading a directory of models at run-time, we propose several key differences for more efficient processing of the input data: 1) we pre-compute additional annotations, e.g. 2D convex hulls, during the build process,
2) natural language queries provide a layer of abstraction that is convenient to interface with neural network outputs, in particular LLMs, 
3) all semantic annotations are directly available from one query,
4) the indexing overhead is shifted from the end user's runtime to our build time,
5) models are released in a semi-obfuscated format, compatible with more licenses than raw files.
\\\noindent
Our default database release contains over 100 obstacles and over 15 human models, fully annotated with over 30 different color-material combinations. Developers can use our open source building pipeline to extend our model repository with their own proprietary model repositories and annotations to generate their own database and bundle. 

\begin{figure}[!htp]
    \centering
    \includegraphics[width=0.99\linewidth]{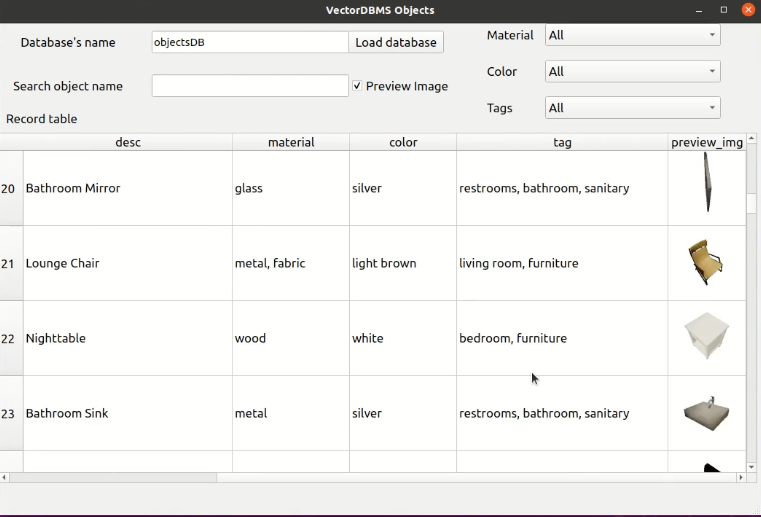}
    \caption{GUI of the Model database. The user is able to integrate their own 3D objects, annotate them, link them semantically with rooms, other objects, or interactions with humans. Objects and assets are queriable by a number of attributes such as color, type, probability to appear in a specific room, their annotations and links to other entities such as rooms or other objects. Using the GUI, the user can specifically set the attributes or include new objects easily.}
    \label{img:dbgui}
\end{figure}

\subsection{ROS2 migration}
\noindent Arena 4.0 is fully operational under ROS2. To achieve this, we developed a number of components specifically for ROS2, including a dynamic map generator and a universal metrics recorder and evaluator. Additionally, we migrated to the Nav2 framework, which provides enhanced navigational functionalities such as recovery behavior, dynamic switching between planners, or enhanced SLAM cababilities out of the box. This migration also simplifies the deployment of the navigation stack on robots, such as quadrupeds, that operate exclusively under ROS2. Furthermore, functionalities such as real-time support, security, and communication protocols are now accessible using ROS2. A table listing all the modules that were migrated from ROS1 to ROS2 is provided in \autoref{tab:modules}.

\begin{table}[!h]
    \centering
    \setlength{\tabcolsep}{1.5pt}
    \renewcommand{\arraystretch}{0.5}
    \caption{Migrated System Components}
    \begin{tabular}{m{2cm}m{2cm}m{4cm}}
        \hline
        \textbf{Arena 3.0} & \textbf{Arena 4.0} & \textbf{Advantage} \\
        \hline
        Ubuntu 20 & Ubuntu 22 & \makecell[l]{modern system, new EOL 2027} \\
        ROS Noetic & ROS2 Humble & \makecell[l]{ROS upgrade, new EOL 2027} \\
        Python3.8 & Python3.10 & \makecell[l]{modern and flexible development,\\ updated GPU libraries} \\
        Rviz & Rviz2 & \makecell[l]{easier to integrate GUI plugins} \\
        2-stage installer & monolithic installer & \makecell[l]{more configuration,\\ respects other workspaces\\ and environments} \\
        system packages & fully-compiled workspace & \makecell[l]{higher system compatibility} \\
        training Docker image & full Docker image & \makecell[l]{development works through\\ WSL, with GUI support} \\
        Gazebo Classic & Gazebo Garden & \makecell[l]{faster simulation,\\ more features} \\
        static imports & lazy imports & \makecell[l]{reduced startup time} \\
        CLI Tools & CLI \& GUI Tools & \makecell[l]{better accessibility} \\
        move\_base\_flex & Nav2 & \makecell[l]{new planner integrations} \\
        ROS1 robots & ROS2 robots & \makecell[l]{access to new robots\\ (e.g. Unitree Go2 and later)} \\
        \hline
    \end{tabular}
    \label{tab:modules}
\end{table}
\begin{figure*}[!h]
    \centering
    \includegraphics[width=0.99\linewidth]{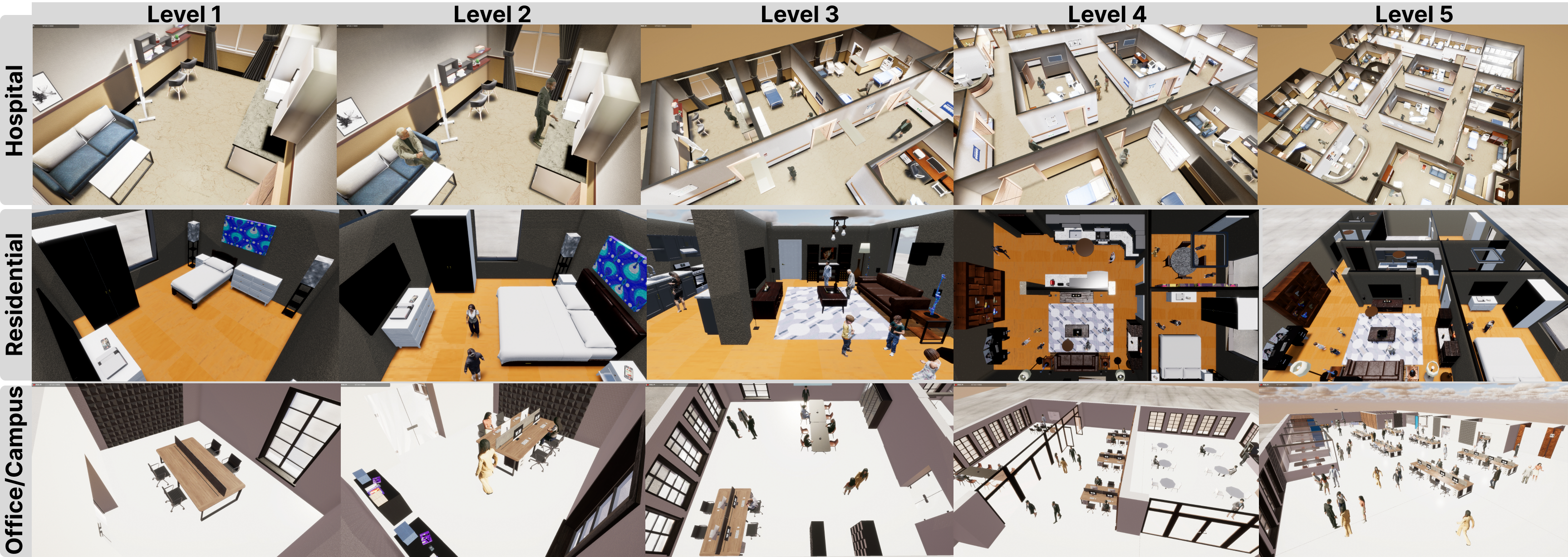}
    \caption{Example worlds generated using the Arena-gen module for benchmarking and competition purposes. The worlds were generated using the text "generate me 5 difficulty levels of a [hospital, residential, office] environment". The assets are automatically taken from the arena model database and pedestrians spawned with HuNavSim. Notably, a large variety of worlds for each environment type and level can be generated, e.g. 500 environments of hospital level 2. This feature aids quantitative benchmarking and in training new models. Users can also customize room layouts, pedestrian interactions with each other, asset placements, and specific situations using the Arena Architect GUI (shown in the supplementary video).}
    \label{img:worlds}
\end{figure*}
\subsection{Scenario Generation and Human Behavior Simulation}
\noindent To extend the human simulations of our previous works, we integrated HuNavSim \cite{perezhunavsim} into our platform and leveraged our new semantic annotations of existing and generated worlds to provide a more complete and realistic simulation of pedestrian behavior. Points are sampled from semantic world regions (\emph{zones}) to produce endless pedestrian behavior scenarios, which are further enhanced by the use of human-human/robot interactions to create a nuanced social environment. Our extensions to HuNavSim allow for more semantic depth, which finds applications in professional environments with distinct pedestrian roles, e.g., hospitals and offices.

\noindent

\subsection{Improvement of Usability Features}
\noindent Several user studies and surveys conducted during our previous works \cite{arena3} provided valuable feedback on the necessity for improvement in usability features such as easy installation and comprehensive documentation. Thus, we have incorporated the responses and feedback to enhance the user experience and development cycle such as the provision of multiple graphical user interfaces for improved user understanding and control customization, as well as editors for annotating training data that directly match the format required to execute functions, e.g., creating new rooms and scenarios. Furthermore, the installation process has been simplified through the use of an installer that guides the user through the installation.

\subsection{Platform for competition and benchmarking}
\noindent Arena 4.0 extends evaluation capabilities by providing a number of functions to enable the platform's use for unified benchmarking. With the provided APIs, we facilitate the integration of methods and new planners. The introduced world generation pipeline is capable of generating a unified set of specialized environments of varying difficulty levels, which is crucial for providing consistent and standardized worlds for testing and benchmarking (examples illustrated in Figure \ref{img:worlds}).
This allows users to test their approaches in specific situations and scenarios, based on individual circumstances and needs. As such, Arena 4.0 is has been selected as the hosting platform for the competition as part of the \href{https://socialnav2025.pages.dev/organization}{\color{blue}SocialNav2025 workshop}.
\begin{figure*}[!h]
    \centering
    \includegraphics[width=0.99\linewidth]{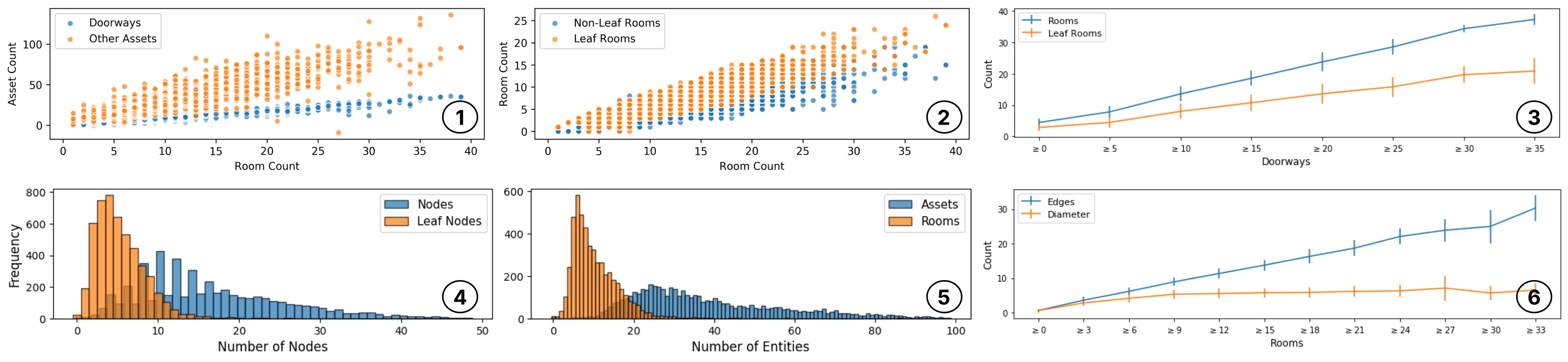}
    \caption{Evaluations of generated worlds based on different room and asset metrics, plotted over input difficulty (1-2) and scene graph metrics (3-6). Following plots are shown: doorways and asset count over the level difficulty (1), Room and Leaf Room count over difficulty (2) and Doorways (3), Frequency of Nodes and Leaf Nodes over the total number of Nodes (4), Nodes and Assets Count over the number of Entities (5), and Edges and Diameters over the number of Rooms (6). The difficulty is estimated by the LLM using the number of rooms and obstacles. Metrics were computed on a dataset consisting of the scene graphs of 5000 generated worlds. The metrics are \emph{(Non-Leaf) Rooms/Nodes}: number of rooms in the world, \emph{Leaf Rooms/Nodes}: number of rooms with only one connection, \emph{Assets}: number of assets placed into rooms, \emph{Edges/Doorways}: number of connections between rooms, \emph{Diameter}: minimum longest (graph) distance between any two nodes, \emph{Difficulty}: a latent, numerical representation of the world complexity.}
    \label{img:eval}
\end{figure*}
\section{Validation and Evaluation}
\noindent
Similar to all our previous works, we conducted a study asking participants to install and test out specific modules of the platform. Subsequently, we evaluated the capabilities of the o module ur proposed world generation module to generate levels of increasing difficulty and variety from text prompts. 

\subsection{User Study}
\noindent
We evaluated our platform through a user study with 20 participants from universities in Germany, Switzerland, the US, Singapore, Vietnam, Japan, and Korea, all with varying levels of robotics expertise. Participants were selected from diverse groups, including students and researchers familiar with previous versions of Arena, social navigation researchers who had never used Arena, new users from our lab, and students from various international universities without prior experience with Arena. Participants were instructed to install and interact with the platform by completing a specific set of tasks and subsequently completed a questionnaire covering the questions about the platform. All questions and responses are  \href{https://drive.google.com/drive/folders/1lPJrvgxsfZyrIH3be6EFNX4R7y3PkL3J?usp=sharing}{\color{blue} publicly available online}. 
\\\noindent
The study results were mostly positive, with participants highlighting the platform's diverse functionalities in world generation and customizable scenario creation as key enhancements. Both new and experienced users emphasized that the simplified installation process greatly improved usability. Most participants expressed interest in using the platform for benchmarking and integrating their own planning algorithms. Some industry and academic participants also suggested that incorporating a more realistic simulator, like Isaac Gym and Habitat, could further enhance the platform. Furthermore, a number of participants found the evaluation and plotting pipeline counterintuitive, preferring the previous version's customizable iPython notebooks over the current web-based interface. We are currently using this feedback to further improve Arena.



\begin{figure}[!h]
    \centering
    \includegraphics[width=0.99\linewidth]{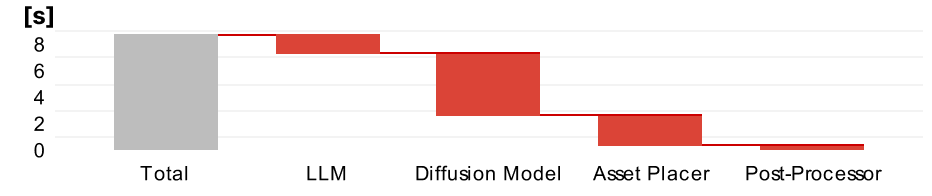}
    \caption{Average latency of each generation component per single generated world. Benchmarks were performed on a single RTX 4080 graphics card and averaged over 100 cold-cache batches. Each individual stage is a separate containerized component and was measured externally by its HTTP response times, adjusted for transmission overhead. The components correspond to \autoref{img:gen-dataflow}.}
    \label{img:evaltime}
\end{figure}

\subsection{World Generation Evaluation}
\noindent
Figure \ref{img:worlds} illustrates examples of worlds generated using Arena-gen at different difficulty levels, based on the prompt: "Generate 50 indoor worlds with 8 difficulty levels.", repeated for 100 batches with different contexts, ranging from residential homes to hospitals. \autoref{img:evaltime} outlines the average computational cost across all worlds. To evaluate the difficulty of the generated levels, we plotted various factors that could potentially contribute to increased difficulty, such as the number of rooms, assets, links to other entities, and the number of pedestrians. \autoref{img:eval} illustrates these plots. We show (1) the asset and doorway count, and (2) the leaf/non-leaf room count, plotted against world difficulty. Thereby, the difficulty is estimated by the LLM using the number of rooms and obstacles. A linear growth of all metrics can be observed with increasing world difficulty. This indicates that real complexity scales well with input difficulty and does not saturate even up to 25-room worlds, covering most practical use cases. The doorway count shows lower variance compared to the other metrics, suggesting a strong correlation between room count and non-leaf room count. (4) and (5) display the absolute frequency of (non-leaf) nodes, leaf nodes, and assets per scene graph in the dataset.
The wide distributions demonstrate the diversity of generated environments. \noindent
(3) shows the mean and variance of the number of rooms and leaf rooms against the number of doorways. A continuous growth of the mean with bounded variance is clearly visible. \noindent
(6) shows the increase in the graph edges count with an increase in the number of rooms, while the graph diameter (minimum longest distance between any two nodes) is kept low. In practice, our generated worlds offer a desirably high graph connectivity, providing robots with several pathways to reach the same goal.


\section{Conclusion}
\noindent 
In this paper, we introduced Arena 4.0, an enhanced version of our previous platforms emphasizing the dynamic generation of 3D environments for training, testing, and benchmarking. Using the Arena models database, users can integrate their own 3D assets and link them semantically to be placed, spawned, and interact with each other. We have fully migrated the platform to operate entirely under ROS2 to ensure compatibility with recent robotics hardware and ease the transition to industry production levels. Furthermore, we improved usability features such as a fast installation process, substantially faster processing times, and simplified usage through the GUIs. Our evaluations and user studies demonstrated substantial improvements in processing efficiency and enhanced user experience. Future work includes the addition of simulators such as Isaac Gym and Habitat, the employment of multi-agent reinforcement learning approaches, as well as automated pipelines for deployment on real robots. Additionally, we aspire to release the second version of Arena-web \cite{kastner2023arenaweb}, which will offer the majority of the current platform's functionalities within a web application.


\addtolength{\textheight}{-1cm} 




\typeout{}
\bibliographystyle{IEEEtran}
\bibliography{main}

\begin{thebibliography}{10}
\providecommand{\url}[1]{#1}
\csname url@samestyle\endcsname
\providecommand{\newblock}{\relax}
\providecommand{\bibinfo}[2]{#2}
\providecommand{\BIBentrySTDinterwordspacing}{\spaceskip=0pt\relax}
\providecommand{\BIBentryALTinterwordstretchfactor}{4}
\providecommand{\BIBentryALTinterwordspacing}{\spaceskip=\fontdimen2\font plus
\BIBentryALTinterwordstretchfactor\fontdimen3\font minus \fontdimen4\font\relax}
\providecommand{\BIBforeignlanguage}[2]{{%
\expandafter\ifx\csname l@#1\endcsname\relax
\typeout{** WARNING: IEEEtran.bst: No hyphenation pattern has been}%
\typeout{** loaded for the language `#1'. Using the pattern for}%
\typeout{** the default language instead.}%
\else
\language=\csname l@#1\endcsname
\fi
#2}}
\providecommand{\BIBdecl}{\relax}
\BIBdecl

\bibitem{arena3}
\BIBentryALTinterwordspacing
L.~K{\"a}stner, V.~Shcherbyna, H.~Zeng, T.~A. Le, M.~H.-K. Schreff, H.~Osmaev, N.~T. Tran, D.~Diaz, J.~Golebiowski, H.~Soh \emph{et~al.}, ``Arena 3.0: Advancing social navigation in collaborative and highly dynamic environments,'' \emph{Robotics Science and Systems}, 2024. [Online]. Available: \url{https://openreview.net/forum?id=BjnICVYhGZ}
\BIBentrySTDinterwordspacing

\bibitem{arena:bench}
\BIBentryALTinterwordspacing
L.~Kastner, T.~Bhuiyan, T.~A. Le, E.~Treis, J.~Cox, B.~Meinardus, J.~Kmiecik, R.~Carstens, D.~Pichel, B.~Fatloun, N.~Khorsandi, and J.~Lambrecht, ``Arena-bench: A benchmarking suite for obstacle avoidance approaches in highly dynamic environments,'' \emph{IEEE Robotics and Automation Letters}, vol.~7, no.~4, p. 9477–9484, Oct. 2022. [Online]. Available: \url{http://dx.doi.org/10.1109/LRA.2022.3190086}
\BIBentrySTDinterwordspacing

\bibitem{arena:rosnav}
L.~K{"a}stner, T.~Buiyan, L.~Jiao, T.~A. Le, X.~Zhao, Z.~Shen, and J.~Lambrecht, ``Arena-rosnav: Towards deployment of deep-reinforcement-learning-based obstacle avoidance into conventional autonomous navigation systems,'' in \emph{2021 IEEE/RSJ International Conference on Intelligent Robots and Systems (IROS)}.\hskip 1em plus 0.5em minus 0.4em\relax IEEE, 2021, pp. 6456--6463.

\bibitem{arena-rosnav-2.0}
L.~Kästner, R.~Carstens, H.~Zeng, J.~Kmiecik, T.~Bhuiyan, N.~Khorsandi, V.~Shcherbyna, and J.~Lambrecht, ``Arena-rosnav 2.0: A development and benchmarking platform for robot navigation in highly dynamic environments,'' 2023.

\bibitem{mirsky2024conflict}
R.~Mirsky, X.~Xiao, J.~Hart, and P.~Stone, ``Conflict avoidance in social navigation—a survey,'' \emph{ACM Transactions on Human-Robot Interaction}, vol.~13, no.~1, pp. 1--36, 2024.

\bibitem{singamaneni2024survey}
P.~T. Singamaneni, P.~Bachiller-Burgos, L.~J. Manso, A.~Garrell, A.~Sanfeliu, A.~Spalanzani, and R.~Alami, ``A survey on socially aware robot navigation: Taxonomy and future challenges,'' \emph{The International Journal of Robotics Research}, p. 02783649241230562, 2024.

\bibitem{francis2023principles}
A.~Francis, C.~P{\'e}rez-d'Arpino, C.~Li, F.~Xia, A.~Alahi, R.~Alami, A.~Bera, A.~Biswas, J.~Biswas, R.~Chandra \emph{et~al.}, ``Principles and guidelines for evaluating social robot navigation algorithms,'' \emph{arXiv preprint arXiv:2306.16740}, 2023.

\bibitem{arena2}
L.~K{\"a}stner, R.~Carstens, H.~Zeng, J.~Kmiecik, T.~Bhuiyan, N.~Khorsandhi, V.~Shcherbyna, and J.~Lambrecht, ``Arena-rosnav 2.0: A development and benchmarking platform for robot navigation in highly dynamic environments,'' in \emph{2023 IEEE/RSJ International Conference on Intelligent Robots and Systems (IROS)}.\hskip 1em plus 0.5em minus 0.4em\relax IEEE, 2023, pp. 11\,257--11\,264.

\bibitem{everett2018motion}
M.~Everett, Y.~F. Chen, and J.~P. How, ``Motion planning among dynamic, decision-making agents with deep reinforcement learning,'' in \emph{2018 IEEE/RSJ International Conference on Intelligent Robots and Systems (IROS)}.\hskip 1em plus 0.5em minus 0.4em\relax IEEE, 2018, pp. 3052--3059.

\bibitem{dugas2020navrep}
D.~Dugas, J.~Nieto, R.~Siegwart, and J.~J. Chung, ``Navrep: Unsupervised representations for reinforcement learning of robot navigation in dynamic human environments,'' in \emph{2021 IEEE International Conference on Robotics and Automation (ICRA)}, 2021, pp. 7829--7835.

\bibitem{chen2019crowd}
C.~Chen, Y.~Liu, S.~Kreiss, and A.~Alahi, ``Crowd-robot interaction: Crowd-aware robot navigation with attention-based deep reinforcement learning,'' in \emph{2019 International Conference on Robotics and Automation (ICRA)}.\hskip 1em plus 0.5em minus 0.4em\relax IEEE, 2019, pp. 6015--6022.

\bibitem{chen2017socially}
Y.~F. Chen, M.~Everett, M.~Liu, and J.~P. How, ``Socially aware motion planning with deep reinforcement learning,'' in \emph{2017 IEEE/RSJ International Conference on Intelligent Robots and Systems (IROS)}.\hskip 1em plus 0.5em minus 0.4em\relax IEEE, 2017, pp. 1343--1350.

\bibitem{song2024socially}
D.~Song, J.~Liang, A.~Payandeh, X.~Xiao, and D.~Manocha, ``Socially aware robot navigation through scoring using vision-language models,'' \emph{arXiv preprint arXiv:2404.00210}, 2024.

\bibitem{sridhar2024nomad}
A.~Sridhar, D.~Shah, C.~Glossop, and S.~Levine, ``Nomad: Goal masked diffusion policies for navigation and exploration,'' in \emph{2024 IEEE International Conference on Robotics and Automation (ICRA)}.\hskip 1em plus 0.5em minus 0.4em\relax IEEE, 2024, pp. 63--70.

\bibitem{hirose2023sacson}
N.~Hirose, D.~Shah, A.~Sridhar, and S.~Levine, ``Sacson: Scalable autonomous control for social navigation,'' \emph{IEEE Robotics and Automation Letters}, 2023.

\bibitem{zu2024language}
W.~Zu, W.~Song, R.~Chen, Z.~Guo, F.~Sun, Z.~Tian, W.~Pan, and J.~Wang, ``Language and sketching: An llm-driven interactive multimodal multitask robot navigation framework,'' in \emph{2024 IEEE International Conference on Robotics and Automation (ICRA)}.\hskip 1em plus 0.5em minus 0.4em\relax IEEE, 2024, pp. 1019--1025.

\bibitem{singamaneni2021human}
P.~T. Singamaneni, A.~Favier, and R.~Alami, ``Human-aware navigation planner for diverse human-robot ineraction contexts,'' in \emph{IEEE/RSJ International Conference on Intelligent Robots and Systems (IROS)}, 2021.

\bibitem{bae2024sit}
J.~W. Bae, J.~Kim, J.~Yun, C.~Kang, J.~Choi, C.~Kim, J.~Lee, J.~Choi, and J.~W. Choi, ``Sit dataset: socially interactive pedestrian trajectory dataset for social navigation robots,'' \emph{Advances in Neural Information Processing Systems}, vol.~36, 2024.

\bibitem{tsoi2020sean}
N.~Tsoi, M.~Hussein, J.~Espinoza, X.~Ruiz, and M.~V{\'a}zquez, ``Sean: Social environment for autonomous navigation,'' in \emph{Proceedings of the 8th International Conference on Human-Agent Interaction}, 2020, pp. 281--283.

\bibitem{tsoi2022sean}
N.~Tsoi, A.~Xiang, P.~Yu, S.~S. Sohn, G.~Schwartz, S.~Ramesh, M.~Hussein, A.~W. Gupta, M.~Kapadia, and M.~V{\'a}zquez, ``Sean 2.0: Formalizing and generating social situations for robot navigation,'' \emph{IEEE Robotics and Automation Letters}, vol.~7, no.~4, pp. 11\,047--11\,054, 2022.

\bibitem{puig2023habitat}
X.~Puig, E.~Undersander, A.~Szot, M.~D. Cote, T.-Y. Yang, R.~Partsey, R.~Desai, A.~W. Clegg, M.~Hlavac, S.~Y. Min \emph{et~al.}, ``Habitat 3.0: A co-habitat for humans, avatars and robots,'' \emph{arXiv preprint arXiv:2310.13724}, 2023.

\bibitem{perezhunavsim}
N.~P{\'e}rez-Higueras, R.~Otero, F.~Caballero, and L.~Merino, ``Hunavsim: A ros 2 human navigation simulator for benchmarking human-aware robot navigation,'' \emph{IEEE Robotics and Automation Letters}, 2023.

\bibitem{sprague2023socialgym}
R.~Chandra, Z.~Sprague, and J.~Biswas, ``Socialgym 2.0: Simulator for multi-robot learning and navigation in shared human spaces,'' in \emph{Proceedings of the AAAI Conference on Artificial Intelligence}, vol.~38, no.~21, 2024, pp. 23\,778--23\,780.

\bibitem{bench_mr}
E.~Heiden, L.~Palmieri, L.~Bruns, K.~O. Arras, G.~S. Sukhatme, and S.~Koenig, ``Bench-mr: A motion planning benchmark for wheeled mobile robots,'' \emph{IEEE Robotics and Automation Letters}, vol.~6, no.~3, pp. 4536--4543, 2021.

\bibitem{robobench}
J.~Weisz, Y.~Huang, F.~Lier, S.~Sethumadhavan, and P.~Allen, ``Robobench: Towards sustainable robotics system benchmarking,'' in \emph{2016 IEEE International Conference on Robotics and Automation (ICRA)}, 2016, pp. 3383--3389.

\bibitem{althoff2017commonroad}
M.~Althoff, M.~Koschi, and S.~Manzinger, ``Commonroad: Composable benchmarks for motion planning on roads,'' in \emph{2017 IEEE Intelligent Vehicles Symposium (IV)}.\hskip 1em plus 0.5em minus 0.4em\relax IEEE, 2017, pp. 719--726.

\bibitem{moll2015benchmarking}
M.~Moll, I.~A. Sucan, and L.~E. Kavraki, ``Benchmarking motion planning algorithms: An extensible infrastructure for analysis and visualization,'' \emph{IEEE Robotics \& Automation Magazine}, vol.~22, no.~3, pp. 96--102, 2015.

\bibitem{mrpb}
J.~Wen, X.~Zhang, Q.~Bi, Z.~Pan, Y.~Feng, J.~Yuan, and Y.~Fang, ``Mrpb 1.0: A unified benchmark for the evaluation of mobile robot local planning approaches,'' in \emph{2021 IEEE International Conference on Robotics and Automation (ICRA)}, 2021, pp. 8238--8244.

\bibitem{nair2022dynabarn}
A.~Nair, F.~Jiang, K.~Hou, Z.~Xu, S.~Li, X.~Xiao, and P.~Stone, ``Dynabarn: Benchmarking metric ground navigation in dynamic environments,'' in \emph{2022 IEEE International Symposium on Safety, Security, and Rescue Robotics (SSRR)}.\hskip 1em plus 0.5em minus 0.4em\relax IEEE, 2022, pp. 347--352.

\bibitem{li2024conav}
C.~Li, X.~Sun, P.~Chen, J.~Fan, Z.~Wang, Y.~Liu, J.~Zhu, C.~Gan, and M.~Tan, ``Conav: A benchmark for human-centered collaborative navigation,'' \emph{arXiv preprint arXiv:2406.02425}, 2024.

\bibitem{armeni20193d}
I.~Armeni, Z.-Y. He, J.~Gwak, A.~R. Zamir, M.~Fischer, J.~Malik, and S.~Savarese, ``3d scene graph: A structure for unified semantics, 3d space, and camera,'' in \emph{Proceedings of the IEEE/CVF international conference on computer vision}, 2019, pp. 5664--5673.

\bibitem{kalervo2019cubicasa5k}
A.~Kalervo, J.~Ylioinas, M.~H{\"a}iki{\"o}, A.~Karhu, and J.~Kannala, ``Cubicasa5k: A dataset and an improved multi-task model for floorplan image analysis,'' in \emph{Image Analysis: 21st Scandinavian Conference, SCIA 2019, Norrk{\"o}ping, Sweden, June 11--13, 2019, Proceedings 21}.\hskip 1em plus 0.5em minus 0.4em\relax Springer, 2019, pp. 28--40.

\bibitem{kastner2023arenaweb}
L.~K{\"a}stner, R.~Carstens, L.~Nahrwold, C.~Liebig, V.~Shcherbyna, S.~Lee, and J.~Lambrecht, ``Demonstrating arena-web: A web-based development and benchmarking platform for autonomous navigation approaches,'' \emph{Robotics: Science and Systems (RSS)}, 2023.

\end{thebibliography}

\end{document}